\documentclass[11pt, final]{article}
\usepackage[final]{template/acl}

\usepackage{times}
\usepackage{latexsym}
\usepackage[T1]{fontenc}
\usepackage[utf8]{inputenc}
\usepackage{microtype}
\usepackage{inconsolata}
\usepackage{graphicx}
\usepackage{amsmath}
\usepackage{booktabs}
\usepackage{tcolorbox}
\usepackage{cleveref}
\usepackage{float}
\usepackage{colortbl}
\usepackage{xcolor}
\usepackage{dblfloatfix}
\usepackage{placeins}
\usepackage{enumitem}
\newtcolorbox[auto counter]{examplebox}[1][]{%
width=\columnwidth,
colback=gray!5!white,
colframe=black!80!white,
boxrule=0.8pt,
left=2pt, right=2pt, top=2pt, bottom=2pt,
fonttitle=\bfseries\small,
title=\small\textbf{Example Rewrites (Box~\thetcbcounter):},
#1
}

\hypersetup{
  linkcolor=black,
  linkbordercolor=white,
  citecolor=black,
  urlcolor=black,
}

\title{Generating Adversarial Texts for Machine Translation via GRPO}

\author{
Florian Zogaj \quad \
Jakob Hütteneder \quad \
Giovanni De Muri \quad \\
\textbf{Federico Villa} \quad \
\textbf{Aryan Sood} \quad \
\textbf{Vilém Zouhar} \\
 ETH Zurich, Switzerland \\
{\tt \{zogajf,jhuetteneder,gdemuri,fvilla,arsood,vzouhar\}@ethz.ch}
}

\date{}

\begin{document}
\maketitle

\begingroup
\renewcommand{\thefootnote}{}
\addtocounter{footnote}{0}
\endgroup

\begin{abstract}
As machine translation (MT) systems continue to improve, standard benchmarks become less informative for exposing remaining weaknesses. Traditional methods for creating challenging test sets rely on expensive manual creation or curation, while automated approaches struggle to produce sets with the necessary translation difficulty and linguistic diversity. We propose a scalable reinforcement-learning-based approach for rewriting existing source texts into instances that are more difficult to translate for MT systems.
We fine-tune a large language model with Group Relative Policy Optimization (GRPO), using reward signals based on translation difficulty together with constraints for semantic similarity, grammaticality, and approximate length preservation. On WMT25, our approach substantially reduces average COMET translation quality from 0.63 to 0.48, while preserving grammaticality and readability, whereas the base model remains at 0.64. Evaluations on the unseen WMT19--WMT24 benchmarks confirm that this behavior generalizes beyond the training data, and human evaluation further shows that the rewrites substantially lower translation quality while incurring a moderate drop in naturalness and only a small change in grammaticality. We release our code to support reproducibility.\footnote{\url{https://github.com/FlorianZogaj/Breaking-MT}}
\end{abstract}

\section{Introduction}
Machine Translation (MT) models have reached a state where standard evaluation benchmarks are no longer sufficiently challenging to expose weaknesses \cite{kocmi-etal-2025-findings,last-translation-benchmark}.
While this strong performance reflects the high quality of modern MT systems, it means that standard tests are less informative for distinguishing model capabilities.
This highlights the urgent need to develop new, challenging test sets to guide future research and deployment decisions.

Traditional methods for constructing challenge sets by manually curating linguistically complex examples that target specific phenomena \cite{amrhein2022aces} are expensive, not scalable, and depend on the predefined design choices.

\begin{figure}[t]
\centerline{\includegraphics[width=0.95\columnwidth]{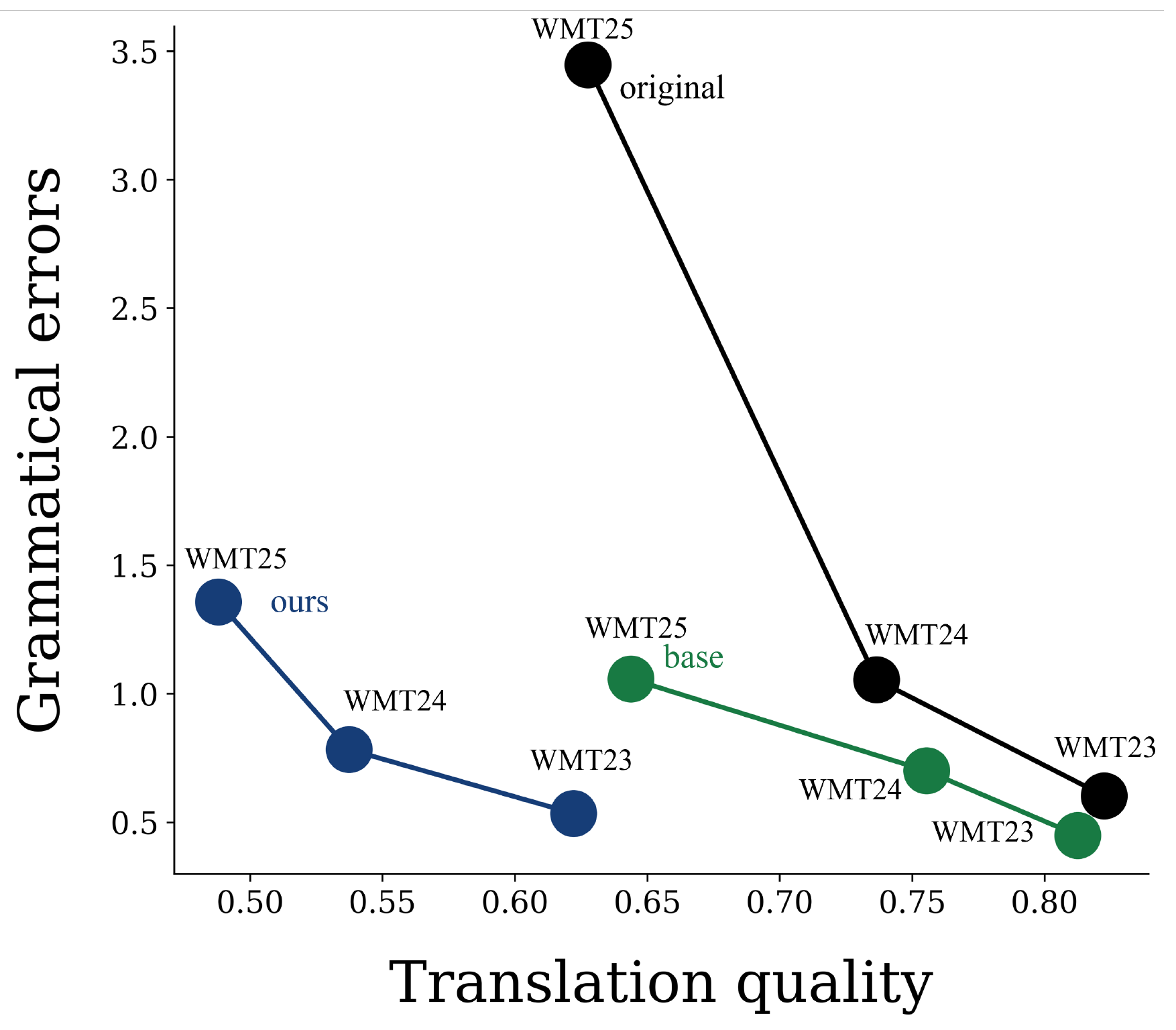}}
\caption{Translation quality (COMET score, scale 0--1; lower is harder to translate) vs. number of grammatical errors on WMT23--WMT25. The rewrites generated by our fine-tuned model (blue) successfully decrease translation quality in comparison to the non-fine-tuned base model (green) and the original dataset (black), while maintaining grammaticality.}
\label{fig:avg_wmt}
\end{figure}

To address this, several automated approaches have been proposed.
For instance, \citet{proietti2025estimatingmachinetranslationdifficulty,kocmi-etal-2025-findings,xu2025searchingdifficulttotranslatetestexamples} analyze natural datasets and filter for difficult-to-translate examples.
However, because naturally occurring sentences are usually simple to translate, this approach is limited and requires massive scale.
Alternatively, \citet{pombal2025zeroshotbenchmarkingframeworkflexible} generate difficult examples from scratch in a zero-shot manner using Large Language Models (LLMs).
While highly scalable, this approach struggles to produce text with sufficient translation difficulty and diversity.

\begin{figure}[t]
\centerline{\includegraphics[width=\columnwidth]{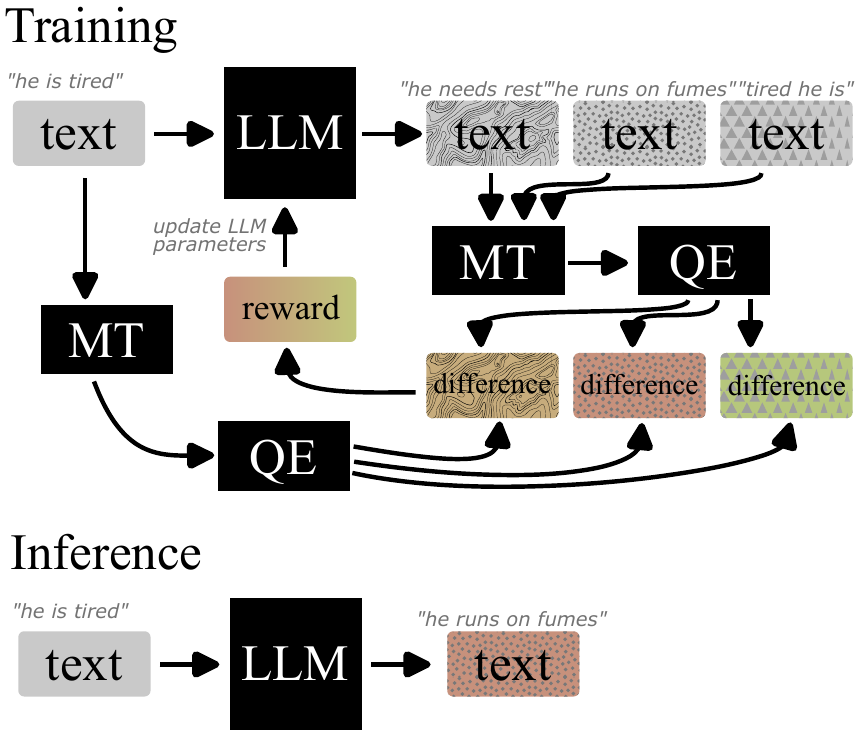}}
\caption{During training, the LLM generates multiple rewrites that are evaluated using machine translation (MT) and quality estimation (QE) models to compute a reward reflected by the difference in translation quality. This signal is used to optimize the model via GRPO. At inference time, the model directly produces rewrites that are difficult to translate. }
\label{fig:pipeline}
\end{figure}

In response to these challenges, we introduce a reinforcement learning-based approach to increase translation difficulty on existing datasets as illustrated in Figure \ref{fig:pipeline}. We fine-tune an LLM using Group Relative Policy Optimization (GRPO, \citealp{shao2024deepseekmathpushinglimitsmathematical}) to automatically rewrite English text into challenging variants for MT systems.
The reward is constructed using translation difficulty estimation models to explicitly encourage modifications that intentionally reduce the expected translation quality.

Figure \ref{fig:avg_wmt} illustrates that our fine-tuned model substantially reduces translation quality on WMT23--WMT25 \cite{kocmi-etal-2025-findings, kocmi-etal-2024-findings, semenov-etal-2023-findings} datasets, while maintaining grammaticality. Further evaluations showing our improvements on WMT datasets \cite{kocmi-etal-2022-findings, akhbardeh-etal-2021-findings, barrault-etal-2020-findings, barrault-etal-2019-findings} can be found in Appendix \ref{app:wmt}.
In addition to automated metrics, we validate the quality of our generations through human evaluation, confirming that the rewrites remain natural while effectively reducing translation quality.

\section{Related Work}
\label{sec:related-work}
\paragraph{Building challenge sets in machine translation.}
Challenge sets are essential in machine translation for probing model weaknesses that are not captured by standard evaluation benchmarks. Traditionally, these sets have been constructed through manual curation or rule-based modifications to target specific linguistic phenomena. For example, ACES \cite{amrhein2022aces} provides a challenge set consisting of 68 distinct linguistic error features. To overcome the scalability limitations, recent research has explored automated methods. One prominent approach is to mine large-scale corpora for challenging sentences that occur naturally. For example, \citet{xu2025searchingdifficulttotranslatetestexamples} formulate the discovery of difficult-to-translate texts as a multi-armed bandit problem. This framework allows them to efficiently search through internet-scale data to identify seed topics that consistently contain examples capable of degrading the performance of machine translation models.

Another direction uses text generation to create challenging examples directly. \citet{zouhar2025generatingdifficulttotranslatetexts} show that LLMs can be used to iteratively modify source sentences to increase their translation difficulty. Inspired by this concept, we instead explicitly train a generator model optimized for this objective, allowing difficult rewrites to be produced directly at inference time.

\paragraph{Adversarial examples with reinforcement learning.}
Reinforcement learning has been used to generate adversarial examples across different settings.
In text classification, \citet{vijayaraghavan2019generating} use Self-Critical Sequence Training to train a model that introduces minimal perturbations to sentences while preserving their semantic content with the goal of fooling a classifier. In the context of machine translation, \citet{zou2020reinforcedgenerationadversarialexamples} use an actor-critic reinforcement learning framework to apply token-level edits that degrade translation quality while preserving the meaning of the original text.
Lastly, \citet{kalikman-etal-2026-augmenting} use signal from a quality estimation in a beam search to increase text translation difficulty.

\paragraph{Translation difficulty.}
Because our primary objective is to generate sentences that are difficult to translate, we need a reliable way to estimate translation difficulty without relying on human references. Prior work suggests two main strategies: generating a translation to explicitly evaluate its quality, and directly estimating the difficulty from the source text alone. We refer to these two approaches as MT+QE and Sentinel, respectively. 

To implement these strategies, we use two distinct models: (1) For the MT+QE pipeline, we use CometKiwi \cite{rei2022cometkiwi}, which predicts translation quality from a source sentence and a candidate machine translation, without requiring a human reference translation and (2) Sentinel \cite{proietti2025estimatingmachinetranslationdifficulty}, which estimates a difficulty score directly from the source sentence alone without requiring either a candidate or a reference translation.
\section{Methods}
\label{method}

\subsection{Generative Model and Prompting}
\paragraph{Generative model.}
We use Llama-3.1-8B-Instruct \cite{grattafiori2024llama} as the rewriting model, since it provides a good balance between instruction-following ability, generation quality, and computational efficiency for reinforcement learning fine-tuning. To make training efficient, we fine-tune it with Low-Rank Adaptation \cite{hu2022lora}.

\paragraph{Prompting.} To guide the model toward generating sentences that are difficult to translate, we include specific instructions and constraints in the prompt along with the input sentence. This prompt explicitly instructs the model to increase translation difficulty of the input text, while preserving its original meaning, grammaticality, and approximate length. The exact prompt is provided in Appendix \ref{app: prompts}.

\subsection{GRPO Fine-Tuning}
We fine-tune the rewriting model using Group Relative Policy Optimization (GRPO). Given an input sentence and the rewriting prompt described above, the model generates multiple candidate rewrites. Each candidate is then assigned a scalar reward that reflects how well it increases translation difficulty while satisfying a set of linguistic constraints. GRPO updates the policy by comparing candidate rewrites within the same prompt group rather than optimizing against a fixed external target. Intuitively, rewrites that receive higher rewards than other candidates for the same source sentence become more likely under the updated policy, while candidates with lower rewards are discouraged. This makes GRPO well suited to our setting, where many possible rewrites may be acceptable, but some achieve a better balance between adversarial strength and linguistic quality than others. In our implementation, the reward is computed separately for each sampled completion and passed to the GRPO trainer as a single weighted score. The full GRPO objective is provided in Appendix \ref{app:grpo}. We use this optimization procedure to train the model to produce rewrites that reduce the expected translation quality while preserving grammaticality and approximate semantics and length.

\subsection{Reward Design}
\label{sec: reward}
Our reward function combines a primary translation difficulty objective with additional constraint rewards that preserve the quality of the rewritten text. For an original sentence $x$ and a generated rewrite $y$, the total reward is defined as
\begin{align}
r(x,y) &=
  \lambda_{\text{td}} \cdot r_{\text{td}}(x,y) \notag\\
&\quad + \lambda_{\text{len}} \cdot r_{\text{len}}(x,y) \notag\\
&\quad + \lambda_{\text{sem}} \cdot r_{\text{sem}}(x,y) \notag\\
&\quad + \lambda_{\text{cola}} \cdot r_{\text{cola}}(y) \notag\\
&\quad + \lambda_{\text{lt}} \cdot r_{\text{lt}}(y),
\end{align}
where $r_{\text{td}}$ is the reward for translation difficulty, $r_{\text{len}}$ encourages the preservation of approximate length, $r_{\text{sem}}$ enforces semantic similarity to the original, and $r_{\text{cola}}$ and $r_{\text{lt}}$ promote grammatical well-formedness. In the reported experiments, we use the weights
$\lambda_{\text{td}} = 1.0$,
$\lambda_{\text{len}} = 0.4$,
$\lambda_{\text{sem}} = 0.8$,
$\lambda_{\text{cola}} = 0.5$, and
$\lambda_{\text{lt}} = 0.2$.

\paragraph{Translation difficulty reward.}
The main training signal is based on the change in the score for translation difficulty between the original sentence and its rewrite. To investigate which formulation is most effective for this objective, we compare two reward strategies using the quality estimation models introduced in Section~\ref{sec:related-work}: (1) MT+QE and (2) Sentinel. Let $\mathrm{QE}(\cdot)$ denote the corresponding difficulty estimator used in a given training run, i.e., either the Sentinel score or a COMET-based score. Since lower scores correspond to more difficult source texts for both difficulty estimation strategies, we define the reward as
\begin{align}
r_{\text{td}}(x,y) = \mathrm{clip}(5 \cdot (\mathrm{QE}(x){-}\mathrm{QE}(y)), -2, 2).
\end{align}
A rewrite therefore receives a positive reward if it obtains a lower quality estimation score than the original sentence. Since the raw difficulty differences are usually small, especially early in training, we scale and clip these values so that they are similar in magnitude to the constraint rewards.

\paragraph{Length reward.}
To discourage the model from increasing difficulty by simply generating long or verbose sentences, we reward rewrites whose length remains close to that of the original input.
Let $\rho = \frac{|y|}{|x|}$ denote the ratio of word counts. If $\rho < 0.5$ or $\rho > 2.0$, we assign a penalty of $-2$. Otherwise, we use a shaped reward that is maximal when $\rho \approx 1$ and decreases as the rewrite becomes shorter or longer than the original.

\paragraph{Semantic similarity reward.}
While semantic faithfulness is not an explicit objective, we use this reward to further discourage the model from resorting to degenerate strategies such as mode collapse or reward hacking. We compute the cosine similarity between sentence embeddings of the original and the rewrite. If the similarity falls below a threshold of $0.70$, the rewrite receives a penalty of $-0.5$, otherwise, it receives a reward of $+0.5$. In earlier versions, we used a shaped reward that increased further as similarity approached $1.0$. However, this encouraged the model to make only minimal edits and empirically led to earlier collapse. We therefore use a formulation that rewards semantic preservation without pushing the model to almost copy the input.

\paragraph{Grammaticality rewards.}
To maintain linguistic plausibility, we enforce grammatical correctness via two independent models. First, we compute a linguistic acceptability score using a RoBERTa-base model fine-tuned on the Corpus of Linguistic Acceptability (CoLA) dataset \cite{warstadt2019neural}. We rescale the predicted acceptability probability to the interval $[-1,1]$. Second, we use LanguageTool to count grammatical errors in the output. The reward is based on the number of errors $e$, defined as $r_{\text{lt}} = 1.0 - 2 \log(1 + e)$, and clipped at $-2$. 
\\\\
The reward weights and reward design hyperparameters were chosen based on preliminary experiments that yielded a stable trade-off between translation difficulty and linguistic quality. Since our primary contribution is demonstrating the effectiveness of GRPO-based adversarial rewriting, we do not focus on exhaustive hyperparameter tuning. A more systematic optimization is left for future work.

\section{Experiments}
\label{sec:experiments}
\paragraph{Setup.}
To thoroughly evaluate the effectiveness and transferability
of our adversarial rewrites, we explore reward signals for translation difficulty from Sentinel and COMET. Since COMET requires a source sentence and translation to compute a score, we additionally ensure that our approach does not overfit to weaknesses of a specific MT model. To this end, we use two translation models: NLLB \cite{costa2022no} and Helsinki \cite{tiedemann2020opus}.
This experimental setup results in four separately fine-tuned models: one optimized for Sentinel, one for NLLB + COMET, one for Helsinki + COMET, and one for the average of the two MT + COMET scores, denoted as COMET-Average in this work. To determine the robustness of these approaches, we conduct a comparative analysis, where we evaluate the outputs of each trained model under all four difficulty metrics.

\paragraph{Language settings.}
During reinforcement learning fine-tuning, translation-difficulty rewards are computed using English$\rightarrow$Italian translations generated by NLLB and Helsinki. Thus, the rewriting model is optimized only against English$\rightarrow$Italian MT performance. In the human evaluation, we instead evaluate English$\rightarrow$German translations generated by NLLB. This allows us to test whether the learned source-side rewriting strategies transfer to a target language that was not used to construct the training reward.

\paragraph{Datasets.}
We train on English source sentences from WMT25, using it as a realistic test case for converting an existing and already difficult MT benchmark into a harder challenge set. Since evaluating only on the training benchmark could reflect dataset-specific adaptation, we additionally discuss evaluation on unseen WMT24, and report WMT19–WMT23 results in Appendix \ref{app:wmt} to assess transfer across earlier benchmarks.

\begin{table*}[!b]
\centering
\small
\setlength{\tabcolsep}{6pt}
\renewcommand{\arraystretch}{1.05}
\begin{tabular}{p{0.31\textwidth} p{0.31\textwidth} p{0.31\textwidth}}
\toprule
\textbf{(a) Original Bigrams} & \textbf{(b) Fine-tuned Bigrams} & \textbf{(c) Fine-tuned Bigrams Collapsed}\\
\midrule
\texttt{"go to"}: 29 & \texttt{"go to"}: 30 & \texttt{"high high"}: \texttt{1965}\\
\texttt{"feel like"}: 18 & \texttt{"low key"}: 28 & 
\texttt{"high stake"}: \texttt{242} \\
\texttt{"look place"}: 18 & \texttt{"real deal"}: 27 &
\texttt{"stake high"}: \texttt{214} \\
\texttt{"price range"}: 18 & \texttt{"dead set"}: 27 &
\texttt{"hook high"}: \texttt{46} \\
\texttt{"x thank"}: 17 & 
\texttt{"get to"}: 23 &
\texttt{"high hook"}: \texttt{41} \\
\bottomrule
\end{tabular}
\caption{Most frequent lemmatized bigrams: (a) original text vs. (b) fine-tuned model at step 2000 vs. (c) collapsed fine-tuned model at step 3500. Primarily optimizing for specific metrics may lead the model to exploit high-reward constructions, resulting in low linguistic diversity (c). Our best performing model (b) is able to increase translation difficulty while maintaining similarly distributed word frequency to the original data (a).}
\label{tab:ngram-lemma-freq}
\end{table*}

\paragraph{What makes a good challenging dataset?}
As there is no standardized way to evaluate dataset quality, we follow a similar evaluation protocol as proposed in \citet{zouhar2025generatingdifficulttotranslatetexts}. For each dataset instance (which may consist of one or more sentences), we compute the following properties and report their averages over the evaluation set:
\begin{itemize}[leftmargin=*, itemsep=-2pt, topsep=2pt]
    \item \textbf{Diversity:} We assess output diversity using pairwise Self-chrF \cite{popovic2015chrf} and the frequency of the most common lemmatized bigrams in order to detect collapse.
    \item \textbf{Grammatical correctness:} We measure the average number of grammatical errors per instance using LanguageTool \cite{language_tool_python}.
    \item \textbf{Entropy:} We compute the mean entropy score following \citet{Hansen2023}.
    \item \textbf{RIX:} We compute the mean RIX readability score following \citet{Hansen2023}.
    \item \textbf{Length statistics:} We report average word length excluding stopwords and average instance length.
\end{itemize}

\noindent
Additionally, we use an LLM-as-judge setup to score several high-level properties (prompt in Appendix \ref{app: prompts}). Specifically, we evaluate:
\begin{itemize}[leftmargin=*, itemsep=-2pt, topsep=2pt]
    \item \textbf{Naturalness:} Likelihood that an instance appears to be human-written.
    \item \textbf{Unique topics:} We extract topics for each instance and report the number of distinct topics across the evaluation set.
    \item \textbf{Syntax complexity:} Average score for how complex and difficult to understand an instance is.
    \item \textbf{Word rarity:} Average rarity of the words used in an instance, based on how often an average modern human would use them.
\end{itemize}

\noindent
For all LLM-based evaluations, we use Llama-3.1-8B-Instruct as the judge model.

\subsection{Results}
Our fine-tuning shows the expected trade-off between reward optimization and output diversity. Early in training, the model learns rewrites that reduce translation quality while remaining fluent and diverse. As optimization continues, the model increasingly exploits high-reward linguistic patterns, reusing constructions that reliably reduce translation quality across many inputs.

This makes checkpoint selection an important part of the method. The best checkpoint is not the one that minimizes translation quality the most, but the one that balances translation difficulty, readability, and diversity. We therefore monitor quality and collapse indicators, including grammaticality, naturalness, and the frequency of common lemmatized bigrams. Table \ref{tab:ngram-lemma-freq} illustrates this behavior by comparing the original data, the selected checkpoint, and a later collapsed checkpoint.

We note that such collapse patterns may also reveal concrete weaknesses of translation models, since the phrases that become overused can point to linguistic phenomena that are handled especially poorly. However, a detailed analysis of these weaknesses is beyond the scope of this paper.

\begin{table*}[t!]
\centering
\resizebox{\textwidth}{!}{%

\begin{tabular}{l cc >{\columncolor{gray!10}}c >{\columncolor{gray!10}}c >{\columncolor{gray!10}}c >{\columncolor{gray!10}}c}
\toprule
\textbf{Metric} & \textbf{Original} & \textbf{Base Model} & \textbf{Sentinel} & \textbf{COMET-NLLB} & \textbf{COMET-Helsinki} & \textbf{COMET-Average} \\
\midrule

\multicolumn{7}{l}{\textit{Diversity}} \\
Unique topics & 403 & 436 & 430 & 406 & 410 & 400 \\
Pairwise Self-chrF & 21.61 & 26.00 & 24.46 & 22.77 & 22.58 & 21.65 \\

\hline
\multicolumn{7}{l}{\textit{Grammatical Correctness}} \\
Grammatical Errors & 3.45 & 1.06 & 0.88 & 1.74 & 1.33 & 1.36 \\

\hline
\multicolumn{7}{l}{\textit{Complexity}} \\
Entropy & 4.23 & 6.09 & 4.98 & 5.28 & 4.86 & 4.39 \\
RIX & 3.51 & 7.36 & 21.60 & 3.39 & 3.31 & 2.53 \\
Average Word Length & 4.54 & 5.15 & 4.92 & 4.42 & 4.48 & 4.26 \\
Average Output Length  & 119.95 & 160.01 & 126.70 & 137.51 & 127.49 & 116.17 \\
Syntax Complexity & 61.35 &  81.24 & 84.26 & 71.60 & 73.22 & 70.04 \\
Word Rarity & 38.21 & 45.17 & 51.43 & 45.10 & 43.68 & 43.71 \\
Naturalness & 81.97 & 91.75 & 87.16 & 84.93 & 88.11 & 87.22 \\

\hline
\multicolumn{7}{l}{\textit{Translation Difficulty Score}} \\
Sentinel & -1.56 & -1.62 & \textbf{-2.01} & -1.97 & -1.91 & -1.95 \\
NLLB + COMET & 0.63 & 0.64 & 0.66 & 0.54 & 0.55 & \textbf{0.51} \\
Helsinki + COMET& 0.62 & 0.65 & 0.65 & 0.50 & 0.50 & \textbf{0.46} \\
Average COMET& 0.63 & 0.64 & 0.66 & 0.52 & 0.52 & \textbf{0.48} \\

\bottomrule
\end{tabular}
}
\caption{Evaluation metrics on WMT25 for the original dataset, the non-fine-tuned base model, and the four GRPO fine-tuned variants (gray). COMET-Average achieves the best scores under all COMET-based evaluations and is competitive on Sentinel, while naturalness and grammaticality improve compared to the original data. All COMET scores in this table are computed for English$\rightarrow$Italian translations.}
\label{tab:wmt25-results}
\end{table*}

\subsection{Analysis of WMT25 Results}
Table \ref{tab:wmt25-results} shows that the fine-tuned models substantially decrease translation quality over both the original dataset and the non-fine-tuned base model, but the trade-offs they achieve are noticeably different. Relative to the original WMT25 sentences, the base model mainly produces cleaner and more elaborate paraphrases: grammatical errors decrease strongly and naturalness increases, but the translation quality metrics remain almost the same. This suggests that following instructions alone is not sufficient to generate challenging adversarial rewrites. Instead, it tends to produce more polished reformulations without making the sentences harder to translate.

Among the fine-tuned variants, the Sentinel-based model performs best only on the Sentinel estimator. While it achieves the lowest Sentinel score, it does not generalize to the COMET-based evaluation metrics, where it even has slightly higher translation quality than the original dataset. In contrast, the COMET-based reward formulations consistently decrease translation quality across all metrics. Both COMET-NLLB and COMET-Helsinki substantially lower the COMET scores while also decreasing the Sentinel score over the original data.

The strongest overall results are obtained by COMET-Average. This model achieves the best average COMET score, as well as the best individual scores under both COMET evaluation settings, while remaining competitive on Sentinel. Together, this makes it the most robust model in Table \ref{tab:wmt25-results}. This result suggests that averaging reward signals across multiple MT systems provides a more stable training objective than optimizing against a single estimator alone.

It is important to note that these gains are not achieved simply by making the text noisier or less grammatical. All fine-tuned models produce fewer grammatical errors than the original dataset, including the best performing COMET-Average. This is an important result, since it shows that the model does not reduce translation quality by just introducing additional errors. At the same time, naturalness remains above the original baseline, although lower than for the non-fine-tuned base model. This aligns with the trade-off discussed at the beginning of this section, since the goal is not to produce the most natural sentences possible, but to find a checkpoint that substantially reduces the translation quality while still generating fluent and readable English.

The complexity-related metrics further support this interpretation. The base model tends to generate much longer and more elaborate outputs, with a clear increase in entropy, RIX, and syntactic complexity. However, this additional verbosity has barely any impact on translation difficulty. In contrast, COMET-Average keeps the average output length close to the original dataset and does not rely on a higher readability complexity. In fact, some of the complexity measures remain close to, or even below, the original values. This suggests that the model is not simply making the sentences longer or more obscure, but rather learning to introduce constructions that are challenging for translation systems.

Regarding the diversity metrics, COMET-Average stays very close to the original data in terms of unique topics and pairwise Self-chrF, which suggests that the selected checkpoint still preserves substantial output diversity. This interpretation is supported by Table \ref{tab:ngram-lemma-freq}, which shows that this checkpoint is not yet in a state where a small number of phrases dominate generation, unlike later checkpoints. The rewrite in Box \ref{box:rewrites} provides an example of how the model produces fluent English while reducing translation quality on all three metrics.

Overall, these findings support the central claim of this paper: GRPO fine-tuning can be used to transform existing datasets into more difficult and informative challenge sets while preserving linguistic quality. The results on WMT25 show that the learned rewrites can substantially reduce translation quality without relying on grammatical degradation or severe collapse. We therefore next evaluate on WMT24 to test whether this behavior transfers beyond the training set and reflects more general weaknesses of MT systems rather than adaptation to dataset-specific patterns. Since COMET-Average provides the best balance between translation difficulty, readability, and diversity, we treat it as our main model in the remainder of the analysis.

\begin{examplebox}[label=box:rewrites]
\footnotesize
\begin{center}
    \textbf{Original} \\ \textit{Sentinel: -0.38} $|$ \textit{NLLB: 0.69} $|$ \textit{Helsinki: 0.49}\\
\end{center}

Meghan's rebranded herself - and it's another valiant attempt to flog poshness to the little people

\vspace{4pt} \hrule 

\begin{center}
    \textbf{Base Model} \\ \textit{Sentinel: -0.29} $|$ \textit{NLLB: 0.85}  $|$ \textit{Helsinki: 0.85}\\
\end{center}

Meghan's made a concerted effort to reposition herself as a high-end lifestyle influencer, but it's another bold attempt to peddle luxury to the masses, who are increasingly skeptical of her attempts to cultivate an air of refinement.
\vspace{4pt} \hrule

\begin{center}
    \textbf{GRPO - COMET-Average (Ours)}\\ \textit{Sentinel:} \textbf{- 1.08} $|$ \textit{NLLB:} \textbf{0.53} $|$ \textit{Helsinki:} \textbf{0.46}
\end{center}

Meghan's gone all high-end hip – and it's another swell shot at slumming it with the common folk
\end{examplebox}

\subsection{Generalization to WMT24}

Table \ref{tab:wmt24results} shows that the COMET-Average model transfers to unseen WMT24. It lowers COMET-Average from 0.74 to 0.54 and Sentinel from -0.54 to -1.21, while keeping grammatical errors below the original and maintaining similar length, Self-chrF, and naturalness. In contrast, the base model leaves COMET scores essentially unchanged. This shows that the learned rewriting behavior is not limited to WMT25-specific patterns.

We further report evaluations on WMT19--WMT23 in Appendix \ref{app:wmt}.

\begin{table}[h!]
\centering
\resizebox{\columnwidth}{!}{%
\begin{tabular}{l cc >{\columncolor{gray!10}}c}
\toprule
\textbf{Metric} & \textbf{Original} & \textbf{Base Model} & \textbf{COMET-Average} \\
\midrule

\multicolumn{4}{l}{\textit{Diversity}} \\
Unique topics & 381 & 468 & 346 \\
Pairwise Self-chrF & 16.51 & 21.19 & 16.85 \\

\hline
\multicolumn{4}{l}{\textit{Grammatical Correctness}} \\
Grammatical Errors & 1.05 & 0.70 & 0.78 \\

\hline
\multicolumn{4}{l}{\textit{Complexity}} \\
Entropy & 1.79 & 2.70 & 1.88 \\
RIX & 3.44 & 8.35 & 2.81 \\
Average Word Length & 4.77 & 5.50 & 4.39 \\
Average Output Length  & 49.14 & 74.32 & 50.21 \\
Syntax Complexity & 47.18 & 73.18 & 55.66 \\
Word Rarity & 31.87 & 42.85 & 41.09 \\
Naturalness & 85.18 & 92.29 & 83.66 \\

\hline
\multicolumn{4}{l}{\textit{Translation Difficulty Score}} \\
Sentinel & -0.54 & -0.84 & \textbf{-1.21} \\
NLLB + COMET & 0.74 & 0.75 & \textbf{0.55} \\
Helsinki + COMET & 0.74 & 0.76 & \textbf{0.53} \\
Average COMET & 0.74 & 0.76 & \textbf{0.54} \\

\bottomrule
\end{tabular}
}
\caption{Evaluation metrics on WMT24, comparing the original dataset, the non-fine-tuned base model, and the COMET-Average fine-tuned model. All COMET scores in this table are computed for English$\rightarrow$Italian translations.}
\label{tab:wmt24results}
\end{table}

\subsection{Training Dynamics and Checkpoint Selection}

\begin{figure*}[h!]
\centerline{\includegraphics[width=0.935\textwidth]{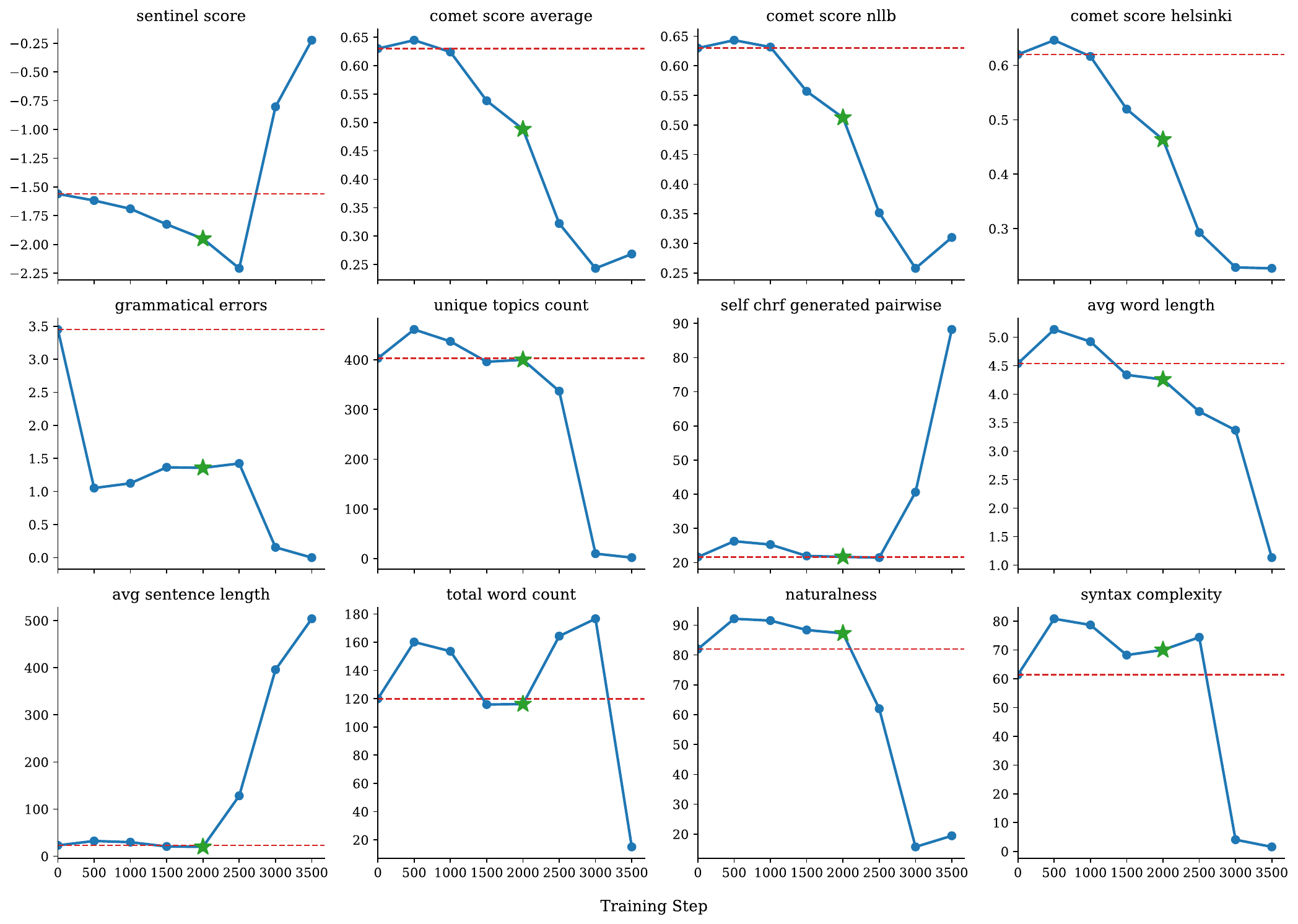}}
\caption{Translation quality, grammaticality, diversity, and complexity metrics change over training steps. The red dashed line corresponds to the original dataset and the green star marks the checkpoint selected for evaluation (step 2000), which reflects the best trade-off between increased difficulty and preservation of linguistic quality.}
\label{fig:checkpoint_eval_all}
\end{figure*}

Figure \ref{fig:checkpoint_eval_all} illustrates why step 2000 was selected as the final checkpoint. In the early and middle stages of training, the model is steadily improving on the main translation quality metrics. At the same time, the metrics related to diversity, grammatical correctness, and complexity remain in a favorable range. Around step 2000, grammatical errors are still low, naturalness remains high, output length is still close to the original scale, and diversity metrics give no indication of collapse.

Earlier checkpoints are less suitable because although they preserve text quality well, they do not yet achieve a substantial reduction in translation quality. In contrast, later checkpoints become increasingly inappropriate despite mostly still improving on the translation quality metrics. Starting after step 2000, the plots show growing instability and clear signs of degeneration: naturalness drops steeply, the number of unique topics decreases strongly, pairwise Self-chrF rises drastically, and sentence length begins to increase to unrealistic levels. In the latest checkpoints, these effects become extreme, indicating that the model is collapsing toward a small set of repeated linguistic constructions.

This interpretation is consistent with the bigram statistics in Table \ref{tab:ngram-lemma-freq}. While the selected checkpoint still shows a varied distribution of common lemmatized bigrams similar to the original data, later checkpoints are dominated by a small number of repeated expressions. For this reason, step 2000 is the most appropriate checkpoint for evaluation: it is the latest point in training at which translation quality has already decreased substantially, while the generations still remain natural, readable, and sufficiently diverse. Qualitative examples of this collapse pattern are provided in Appendix \ref{app:collapse}.

\subsection{Human Evaluation}
\begin{figure}[h!]
\centering
\begin{tikzpicture}
\node[inner sep=0, rounded corners=5pt, clip] {
    \includegraphics[width=\columnwidth]{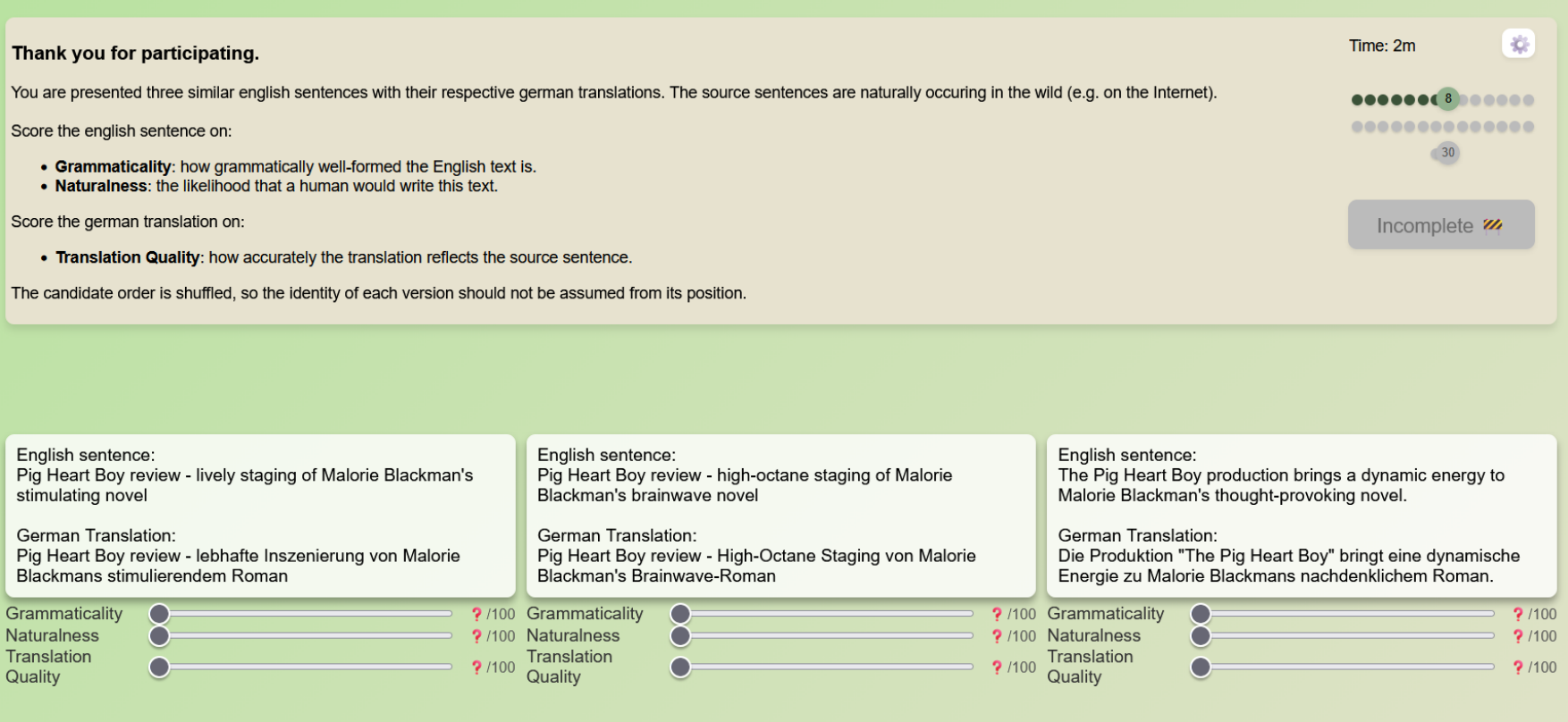}
};
\end{tikzpicture}
\caption{Modified Pearmut annotation interface for text and translation quality annotation.}
\label{fig:pearmut}
\end{figure}

To complement our automated metrics and assess the quality of our generated sentences, we conduct a human evaluation study on WMT25 instances using Pearmut \cite{zouhar2026pearmut} as shown in Figure \ref{fig:pearmut}. In each annotation instance, annotators were presented with three English sentences: the original source sentence, a rewrite produced by the base model, and a rewrite produced by our fine-tuned model. Each English sentence was shown together with its translation into German, generated by NLLB. 
Unlike training, which uses English$\rightarrow$Italian translation signals, the human evaluation used English$\rightarrow$German translations. German was therefore unseen during optimization, making the human evaluation a test of cross-target-language transfer.
The three variants within one instance were randomly shuffled so that annotators could not infer the identity of a sentence from its position.

\paragraph{Task.}
For each of the three versions in a given segment, annotators were asked to use a continuous slider on a 1-100 scale to rate:\\
1. \textbf{Grammaticality:} How grammatically well-formed the English source text is. \\
 2. \textbf{Naturalness:} The likelihood that a human would write this text.\\
 3. \textbf{Translation Quality:} How accurately the target language translation reflects the source sentence.

 \paragraph{Annotators.}
 We collected annotations from 10 annotators. All annotators were native speakers of German and had English proficiency at CEFR level C1 or higher. Each of them evaluated 30 segments, where one segment consists of the original sentence and the two corresponding rewrites. This gives us 300 annotations per condition in total (original, base model, and fine-tuned model). We evaluated 50 English source segments from the WMT25 general MT test set. Candidates were drawn from the pool of unique WMT25 English source texts longer than 10 words. Each list was an independent uniform random sample of 30 of the 50 segments (without replacement within a list). Each of the 10 annotators completed one list.\\

\begin{table}[ht!]
\centering
\resizebox{\columnwidth}{!}{
\begin{tabular}{lccc}
\toprule
\textbf{} & \textbf{Grammaticality} & \textbf{Naturalness} & 
\begin{tabular}{@{}c@{}}\textbf{Translation}\\\textbf{Quality}\end{tabular} \\
\midrule
Original   & 69.75 & \textbf{67.10} & 56.51 \\
(Base Model) & \textbf{74.79} & 64.43 & 57.56 \\
Ours       & 67.13 & 60.01 & \textbf{41.29} \\
\bottomrule
\end{tabular}
}
\caption{Human evaluation results on a 1--100 scale. Our fine-tuned
model achieves substantially lower translation quality, accompanied
by a moderate drop in naturalness and only a small change in grammaticality.}
\label{tab:human_eval}
\end{table}
As shown in Table~\ref{tab:human_eval}, our fine-tuned model
substantially decreases human-rated translation quality, from 56.51 for
the original sources to 41.29 for our rewrites. In contrast, the
base-model rewrites receive a similar score of 57.56.

The human judgments broadly support the trade-off observed in the
automatic evaluation, but reveal a discrepancy in source-text quality.
While the automatic metrics indicate similarly strong or slightly
improved grammaticality and naturalness, annotators rate our rewrites
as somewhat less natural and grammatical. Human raters may be more sensitive to
nested clauses and idiomatic or stylistically unusual phrasing that is
not captured by explicit grammar errors or LLM-based scalar judgments.
We therefore treat the LLM-based naturalness and complexity scores as
diagnostic measures rather than calibrated estimates of human judgment. Moreover, since training used only
English$\rightarrow$Italian translation signals and the human evaluation was conducted on English$\rightarrow$German translations, this also provides
preliminary evidence that the learned rewriting strategies transfer to
an unseen target language.

\begin{figure}[h]
  \centering
  \includegraphics[width=\columnwidth]{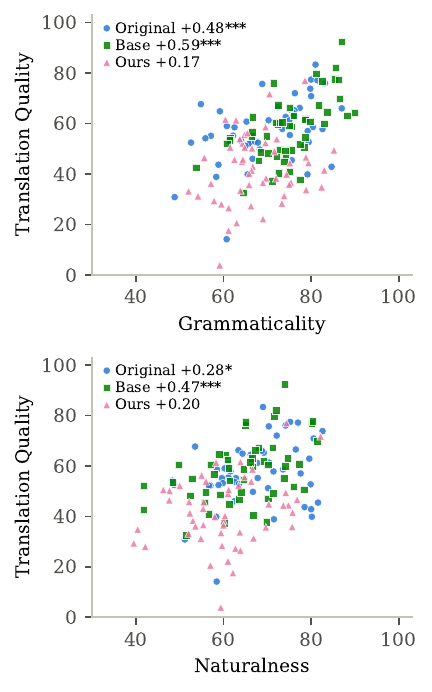}
  \caption{Human-rated translation quality against grammaticality (top)
and naturalness (bottom). Each point represents one source segment
averaged over its annotations ($n{=}50$ per condition). Legends report
within-condition Spearman correlations
($^{*}\,p<.05$, $^{***}\,p<.001$, two-sided). 
Translation quality is not significantly associated with either
dimension in our condition.}
  \label{fig:human-corr}
\end{figure}

\paragraph{Is the drop explained by reduced fluency?}
A natural concern is that our rewrites are harder to translate only because they are less fluent. However, the 15.2 point reduction in
translation quality is much larger than the reductions in naturalness
(7.1 points) and grammaticality (2.6 points). Moreover,
Figure~\ref{fig:human-corr} shows that translation quality correlates
with grammaticality in the original and base-model conditions
($\rho_s=0.48$ and $0.59$), but not with grammaticality or naturalness
in ours ($\rho_s=0.17$ and $0.20$, both $p>0.15$). Therefore, the rewrites
rated hardest to translate are not systematically those rated least
fluent, suggesting that reduced fluency alone does not explain the
effect.

\section{Conclusion}
In this work, we proposed a scalable GRPO-based approach for rewriting existing datasets into more difficult-to-translate variants. The model is optimized to reduce translation quality while preserving naturalness, grammaticality, and approximate meaning. 

Our experiments show that GRPO fine-tuning substantially lowers translation quality without simply producing noisy or ungrammatical text. On WMT25, COMET-Average provides the strongest trade-off between adversarial strength, readability, and diversity. Results on WMT19--WMT24 show that the effect transfers beyond the training data, and human evaluation confirms that the rewrites substantially reduce translation quality with only a moderate drop in grammaticality and naturalness. 

Overall, this suggests that reinforcement learning-based adversarial rewriting can become a practical component of MT evaluation, helping reveal weaknesses that standard benchmarks increasingly fail to capture.

\newpage

\section*{Limitations}
Our approach requires reinforcement learning fine-tuning, which is computationally more expensive and complex than zero-shot generation or filtering-based methods. However, this is a one-time cost that can be amortized across future datasets, since the trained model can then generate adversarial rewrites directly at inference time. 

The experiments rely on difficulty estimation models, which may be exploited by the generator and do not perfectly reflect human translation judgments. In particular, the LLM-based diagnostic scores are not calibrated measures, and human evaluation indicates that they can overestimate naturalness.

Finally, COMET-based training rewards and COMET-based automatic evaluations are limited to English→Italian translations, while cross-target-language transfer is evaluated only on English$\rightarrow$German. Although the human results provide preliminary evidence that the learned rewriting strategies transfer to an unseen target language, they do not establish generalization across a broader range of language pairs, particularly low-resource or typologically distant languages.

\bibliography{refs}

\newpage
\appendix

\section{Prompts}
\label{app: prompts}
\begin{tcolorbox}[
    width=\columnwidth,
    colback=gray!5!white,
    colframe=black!80!white,
    title=\small \textbf{Generation Prompt},
    fonttitle=\bfseries\small,
    boxrule=0.8pt,
    left=2pt, right=2pt, top=2pt, bottom=2pt
]
\footnotesize
You are a rewriting assistant.
Rewrite the given English sentence(s) so that it becomes harder to translate into other languages,
while preserving the original meaning as much as possible.

Guidelines:\\
- Keep the rewrite fluent, grammatical English (no gibberish).\\
- You may increase translation difficulty by using idioms, phrasal verbs, wordplay, ambiguity,
  nested clauses, unusual but natural collocations, or culturally-bound expressions.\\
- Do NOT add explanations, commentary, or multiple options.\\
- Keep length roughly similar: between half and twice the original length.\\

Only output the rewritten sentence(s), no comments or additional text.
Sentence(s) you have to change: \{sentence\}
\end{tcolorbox}

\begin{tcolorbox}[
    width=\columnwidth,
    colback=gray!5!white,
    colframe=black!80!white,
    title=\small \textbf{LLM-as-Judge Prompt},
    fonttitle=\bfseries\small,
    boxrule=0.8pt,
    left=2pt, right=2pt, top=2pt, bottom=2pt
]
\footnotesize
    Analyze the following text and return the answer in JSON.\\
    We want to determine the following attributes:\\
    - naturalness: on a scale from 0 (wholly unnatural) to 100 (fully human-like).\\
    - word rarity: on average, how rare are the words from 0 (used daily) to 100 (unknown).\\
    - syntax complexity: on a scale from 0 (simplest) to 100 (most complex).\\
    - topics: list of 1 to 5 topics that the text is about.\\
    Provide only the output in JSON and nothing else.\\
    The output should look like this: \{\{"naturalness": 80, "word rarity": 50,
    syntax complexity": 70, "topics": ["science", "technology"]\}\}\\
    The sentence to analyze is: \{source\_text\}
\end{tcolorbox}

\newpage
\section{Collapse Pattern}
\label{app:collapse}

Box \ref{box:rewrites-over-time} provides a concrete example of the collapse pattern discussed above. In the earlier checkpoints, the model produces increasingly challenging but still fluent and semantically interpretable rewrites. After step 2000, the outputs begin to converge toward a narrow set of preferred constructions. In this example, the model starts to overuse expressions involving \textit{high}, first generating phrases such as \textit{high-five to that} at step 2500 and then collapsing almost entirely to repeated \textit{high-five} at step 3000. The same pattern is visible in Table \ref{tab:ngram-lemma-freq}, where the collapsed checkpoint is dominated by bigrams containing \textit{high}. 

\begin{examplebox}[label=box:rewrites-over-time]
\footnotesize
\begin{center}
    \textbf{Original} \\ \textit{Sentinel: 0.09} $|$ \textit{NLLB: 0.67} $|$ \textit{Helsinki: 0.47}\\
\end{center}

This will be the year that Guinness loses its cool. Cheers to that!

\vspace{4pt} \hrule 

\begin{center}
    \textbf{COMET-Average step 500} \\ \textit{Sentinel: -0.82} $|$ \textit{NLLB: 0.62}  $|$ \textit{Helsinki: 0.67}\\
\end{center}

This is the year that the iconic stout's mystique is bound to lose its edge, and I'm raising a glass to that impending shift in the cultural zeitgeist.
\vspace{4pt} \hrule

\begin{center}
    \textbf{COMET-Average step 1000}\\ \textit{Sentinel: -0.81} $|$ \textit{NLLB: 0.56} $|$ \textit{Helsinki: 0.64}
\end{center}

This is the year that the iconic stout's mystique starts to lose its fizz, and its reputation begins to go flat. May the bubble finally burst on its pedestal.
\vspace{4pt} \hrule
\begin{center}
    \textbf{COMET-Average step 1500}\\ \textit{Sentinel: -0.29} $|$ \textit{NLLB: 0.62} $|$ \textit{Helsinki: 0.55}
\end{center}

This is the year that the Guinness mystique goes flat. Bottoms up to that!

\vspace{4pt} \hrule
\begin{center}
    \textbf{COMET-Average step 2000}\\ \textit{Sentinel: -0.95}  $|$ \textit{NLLB: 0.40} $|$ \textit{Helsinki: 0.59}
\end{center}
This'll be the year Guinness gets its mojo whacked. Raise a glass to that!

\vspace{4pt} \hrule
\begin{center}
    \textbf{COMET-Average step 2500}\\ \textit{Sentinel: -0.89}$|$ \textit{NLLB: 0.40} $|$ \textit{Helsinki: 0.31}
\end{center}
Bust a gut – this lowdown's the year Guinness goes flat – high-five to that!
\vspace{4pt} \hrule
\begin{center}
    \textbf{COMET-Average step 3000}\\ \textit{Sentinel: -0.17} $|$ \textit{NLLB:0.30} $|$ \textit{Helsinki: 0.23}
\end{center}
High-five – high-five – high-five – high-five – high-five – high-five – – – – – – – –
\end{examplebox}

\onecolumn
\section{GRPO Objective}
\label{app:grpo}

For completeness, we provide the GRPO objective used during fine-tuning.
Let $\pi_\theta$ denote the current policy and $\pi_{\theta_{\mathrm{old}}}$ the reference policy.
For each prompt $q$, we sample a group of $G$ completions
\[
\{o_i\}_{i=1}^G = \{o_1, o_2, \dots, o_G\}
\]
from the old policy. These samples are kept fixed for $L$ optimization steps, during which the current policy is updated by maximizing

\begin{align}
\mathcal{L}(\pi_\theta, \pi_{\theta_{\mathrm{old}}}, \{o_i\}_{i=1}^G)
=
\frac{1}{G}\sum_{i=1}^{G}\frac{1}{|o_i|}\sum_{t=1}^{|o_i|}
l_{i,t}
-\beta D_{\mathrm{KL}}\!\left(\pi_\theta \,\|\, \pi_{\theta_{\mathrm{old}}}\right),
\end{align}

where

\begin{align}
l_{i,t}
&=
\min \Bigl(
s_{i,t}(\theta) A_i,\,
\mathrm{clip}(s_{i,t}(\theta), 1-\varepsilon, 1+\varepsilon) A_i
\Bigr), \\
s_{i,t}(\theta)
&=
\frac{
\pi_\theta(o_{i,t}\mid q, o_{i,<t})
}{
\pi_{\theta_{\mathrm{old}}}(o_{i,t}\mid q, o_{i,<t})
}, \\
A_i
&=
\frac{
r_i - \mathrm{mean}(\{r_1,\dots,r_G\})
}{
\mathrm{std}(\{r_1,\dots,r_G\})
}.
\end{align}
Here, $r_i$ denotes the scalar reward assigned to completion $o_i$, as defined in Section \ref{sec: reward}. Intuitively, GRPO increases the probability of completions that achieve higher reward relative to other samples from the same prompt, while discouraging completions with lower reward.

\twocolumn

\section{Evaluation metrics on WMT19--WMT23}
\label{app:wmt}

This section reports the same automatic evaluation metrics as in the main paper for WMT19--WMT23. For each year, we compare the original dataset, the non-fine-tuned base model, and the COMET-Average model at the selected checkpoint to assess whether the observed trade-off between translation difficulty and text quality also holds across earlier benchmark years. All COMET scores are computed for English$\rightarrow$Italian translations.

\begin{table}[H]
\centering
\resizebox{\columnwidth}{!}{%
\begin{tabular}{l cc >{\columncolor{gray!10}}c}
\toprule
\textbf{Metric} & \textbf{Original} & \textbf{Base Model} & \textbf{COMET-Average} \\
\midrule

\multicolumn{4}{l}{\textit{Diversity}} \\
Unique topics & 266 & 353 & 246 \\
Pairwise Self-chrF & 16.91 & 21.54 & 16.21 \\

\hline
\multicolumn{4}{l}{\textit{Grammatical Correctness}} \\
Grammatical Errors & 0.75 & 0.67 & 0.62 \\

\hline
\multicolumn{4}{l}{\textit{Complexity}} \\
Entropy & 0.83 & 1.39 & 0.86 \\
RIX & 6.05 & 11.54 & 4.22 \\
Average Word Length & 5.44 & 6.14 & 4.84 \\
Average Output Length  & 24.93 & 41.48 & 25.19 \\
Syntax Complexity & 40.52 & 69.84 & 43.84 \\
Word Rarity & 26.33 & 39.49 & 37.47 \\
Naturalness & 87.82 & 93.65 & 80.49 \\

\hline
\multicolumn{4}{l}{\textit{Translation Difficulty Score}} \\
Sentinel & -0.02 & -0.40 & \textbf{-0.55} \\
NLLB + COMET & 0.86 & 0.83 & \textbf{0.68} \\
Helsinki + COMET& 0.83 & 0.81 & \textbf{0.59} \\
Average COMET & 0.84 & 0.82 & \textbf{0.63} \\

\bottomrule
\end{tabular}
}
\caption{Evaluation metrics on WMT19}
\end{table}

\begin{table}[H]
\centering
\resizebox{\columnwidth}{!}{%
\begin{tabular}{l cc >{\columncolor{gray!10}}c}
\toprule
\textbf{Metric} & \textbf{Original} & \textbf{Base Model} & \textbf{COMET-Average} \\
\midrule

\multicolumn{4}{l}{\textit{Diversity}} \\
Unique topics & 283 & 380 & 267 \\
Pairwise Self-chrF & 16.95 & 22.15 & 16.51 \\

\hline
\multicolumn{4}{l}{\textit{Grammatical Correctness}} \\
Grammatical Errors & 0.76 & 0.62 & 0.63 \\

\hline
\multicolumn{4}{l}{\textit{Complexity}} \\
Entropy & 0.91 & 1.56 & 0.97 \\
RIX & 5.82 & 11.92 & 3.95 \\
Average Word Length & 5.46 & 6.24 & 4.79 \\
Average Output Length  & 27.40 & 46.85 & 28.21 \\
Syntax Complexity & 42.73 & 74.43 & 47.37 \\
Word Rarity & 28.05 & 42.84 & 39.49 \\
Naturalness & 85.65 & 93.33 & 80.31 \\

\hline
\multicolumn{4}{l}{\textit{Translation Difficulty Score}} \\
Sentinel & -0.08 & -0.39 & \textbf{-0.69} \\
NLLB + COMET & 0.83 & 0.83 & \textbf{0.64} \\
Helsinki + COMET & 0.83 & 0.82 & \textbf{0.58} \\
Average COMET & 0.83 & 0.83 & \textbf{0.61} \\

\bottomrule
\end{tabular}
}
\caption{Evaluation metrics on WMT20}
\end{table}

\begin{table}[H]
\centering
\resizebox{\columnwidth}{!}{%
\begin{tabular}{l cc >{\columncolor{gray!10}}c}
\toprule
\textbf{Metric} & \textbf{Original} & \textbf{Base Model} & \textbf{COMET-Average} \\
\midrule

\multicolumn{4}{l}{\textit{Diversity}} \\
Unique topics & 269 & 350 & 228 \\
Pairwise Self-chrF & 17.24 & 22.37 & 16.27 \\

\hline
\multicolumn{4}{l}{\textit{Grammatical Correctness}} \\
Grammatical Errors & 0.90 & 0.75 & 0.81 \\

\hline
\multicolumn{4}{l}{\textit{Complexity}} \\
Entropy & 0.86 & 1.41 & 0.91 \\
RIX & 6.41 & 12.78 & 4.71 \\
Average Word Length & 5.48 & 6.17 & 4.83 \\
Average Output Length  & 25.90 & 42.30 & 26.57 \\
Syntax Complexity & 40.76 & 69.66 & 45.31 \\
Word Rarity & 26.51 & 40.93 & 41.20 \\
Naturalness & 89.09 & 93.84 & 78.21 \\

\hline
\multicolumn{4}{l}{\textit{Translation Difficulty Score}} \\
Sentinel & -0.11 & -0.44 & \textbf{-0.68} \\
NLLB + COMET & 0.86 & 0.84 & \textbf{0.67} \\
Helsinki + COMET & 0.84 & 0.82 & \textbf{0.59} \\
Average COMET & 0.85 & 0.83 & \textbf{0.63} \\

\bottomrule
\end{tabular}
}
\caption{Evaluation metrics on WMT21}
\end{table}

\begin{table}[H]
\centering
\resizebox{\columnwidth}{!}{%
\begin{tabular}{l cc >{\columncolor{gray!10}}c}
\toprule
\textbf{Metric} & \textbf{Original} & \textbf{Base Model} & \textbf{COMET-Average} \\
\midrule

\multicolumn{4}{l}{\textit{Diversity}} \\
Unique topics & 375 & 470 & 320 \\
Pairwise Self-chrF & 16.23 & 21.36 & 15.64 \\

\hline
\multicolumn{4}{l}{\textit{Grammatical Correctness}} \\
Grammatical Errors & 0.51 & 0.36 & 0.39 \\

\hline
\multicolumn{4}{l}{\textit{Complexity}} \\
Entropy & 0.81 & 1.41 & 0.85 \\
RIX & 5.52 & 11.36 & 3.59 \\
Average Word Length & 5.19 & 5.90 & 4.53 \\
Average Output Length  & 24.32 & 41.65 & 24.35 \\
Syntax Complexity & 38.61 & 66.64 & 41.03 \\
Word Rarity & 28.85 & 37.51 & 36.70 \\
Naturalness & 85.57 & 92.65 & 80.11 \\

\hline
\multicolumn{4}{l}{\textit{Translation Difficulty Score}} \\
Sentinel & 0.01 & -0.32 & \textbf{-0.49} \\
NLLB + COMET & 0.85 & 0.83 & \textbf{0.66} \\
Helsinki + COMET & 0.83 & 0.81 & \textbf{0.60} \\
Average COMET & 0.84 & 0.82 & \textbf{0.63} \\

\bottomrule
\end{tabular}
}
\caption{Evaluation metrics on WMT22}
\end{table}

\begin{table}[H]
\centering
\resizebox{\columnwidth}{!}{%
\begin{tabular}{l cc >{\columncolor{gray!10}}c}
\toprule
\textbf{Metric} & \textbf{Original} & \textbf{Base Model} & \textbf{COMET-Average} \\
\midrule

\multicolumn{4}{l}{\textit{Diversity}} \\
Unique topics & 359 & 425 & 312 \\
Pairwise Self-chrF & 16.54 & 21.24 & 15.90 \\

\hline
\multicolumn{4}{l}{\textit{Grammatical Correctness}} \\
Grammatical Errors & 0.60 & 0.43 & 0.52 \\

\hline
\multicolumn{4}{l}{\textit{Complexity}} \\
Entropy & 1.07 & 1.63 & 1.09 \\
RIX & 5.81 & 11.40 & 4.03 \\
Average Word Length & 5.29 & 6.03 & 4.71 \\
Average Output Length  & 30.31 & 47.33 & 30.11 \\
Syntax Complexity & 43.61 & 70.66 & 46.70 \\
Word Rarity & 30.29 & 40.19 & 40.76 \\
Naturalness & 84.75 & 93.32 & 80.18 \\

\hline
\multicolumn{4}{l}{\textit{Translation Difficulty Score}} \\
Sentinel & -0.16 & -0.43 & \textbf{-0.72} \\
NLLB + COMET & 0.83 & 0.83 & \textbf{0.65} \\
Helsinki + COMET & 0.82 & 0.80 & \textbf{0.58} \\
Average COMET & 0.82 & 0.81 & \textbf{0.62} \\

\bottomrule
\end{tabular}
}
\caption{Evaluation metrics on WMT23}
\end{table}

\section{Model Evaluations on WMT25}
This section shows how translation difficulty evolves during training on WMT25 for each fine-tuned model. Each figure corresponds to one reward configuration and reports all four difficulty metrics. COMET scores are shown on the left $y$-axis and Sentinel scores on the right $y$-axis. The dashed vertical line marks the checkpoint selected for the evaluation in the main paper.

\begin{figure}[H]
\centerline{\includegraphics[width=\columnwidth]{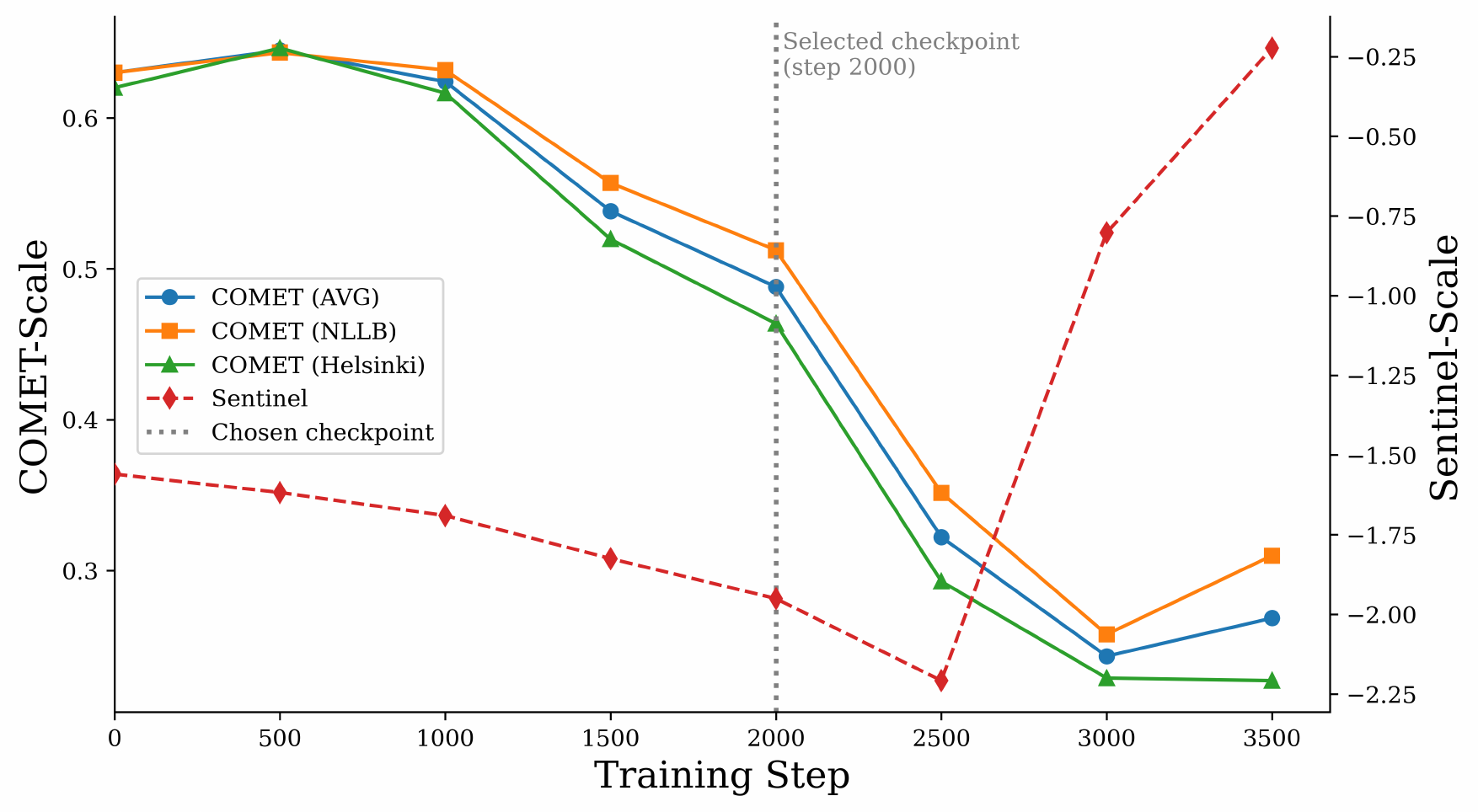}}
\caption{WMT25: Evolution of translation difficulty over training using COMET-Average.}
\end{figure}

\begin{figure}[H]
\centerline{\includegraphics[width=\columnwidth]{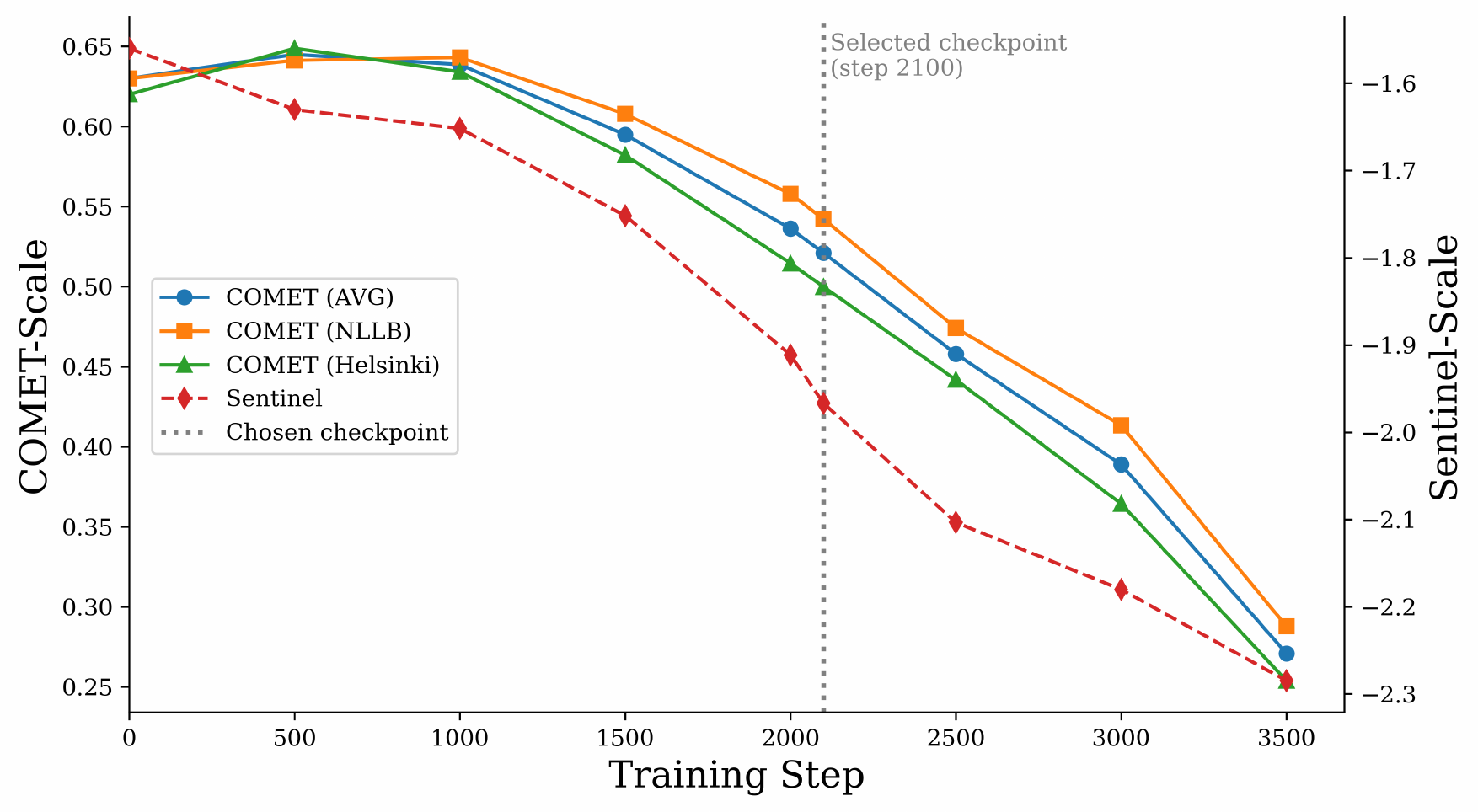}}
\caption{WMT25: Evolution of translation difficulty over training using COMET-NLLB.}
\end{figure}

\begin{figure}[H]
\centerline{\includegraphics[width=\columnwidth]{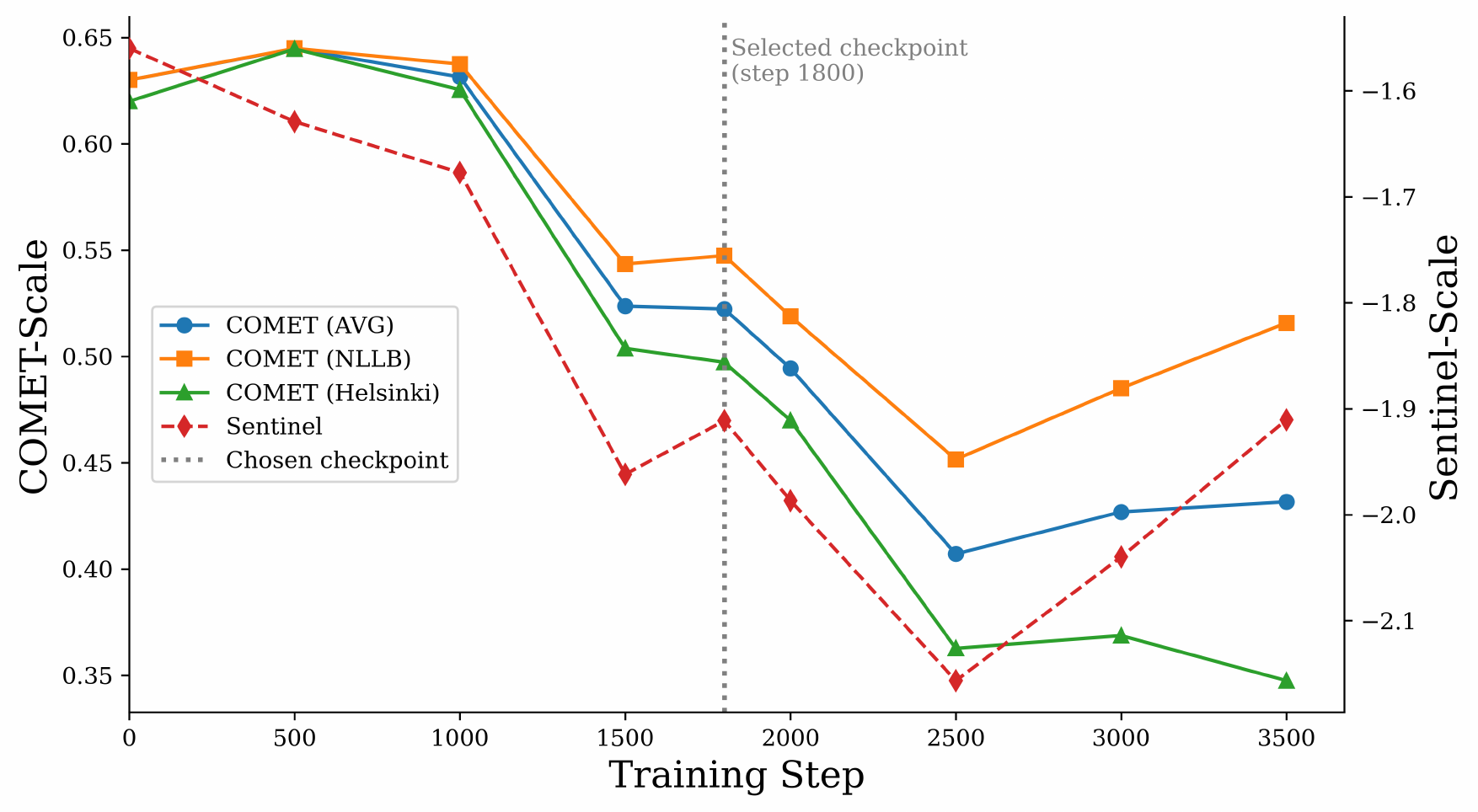}}
\caption{WMT25: Evolution of translation difficulty over training using COMET-Helsinki.}
\end{figure}

\begin{figure}[H]
\centerline{\includegraphics[width=\columnwidth]{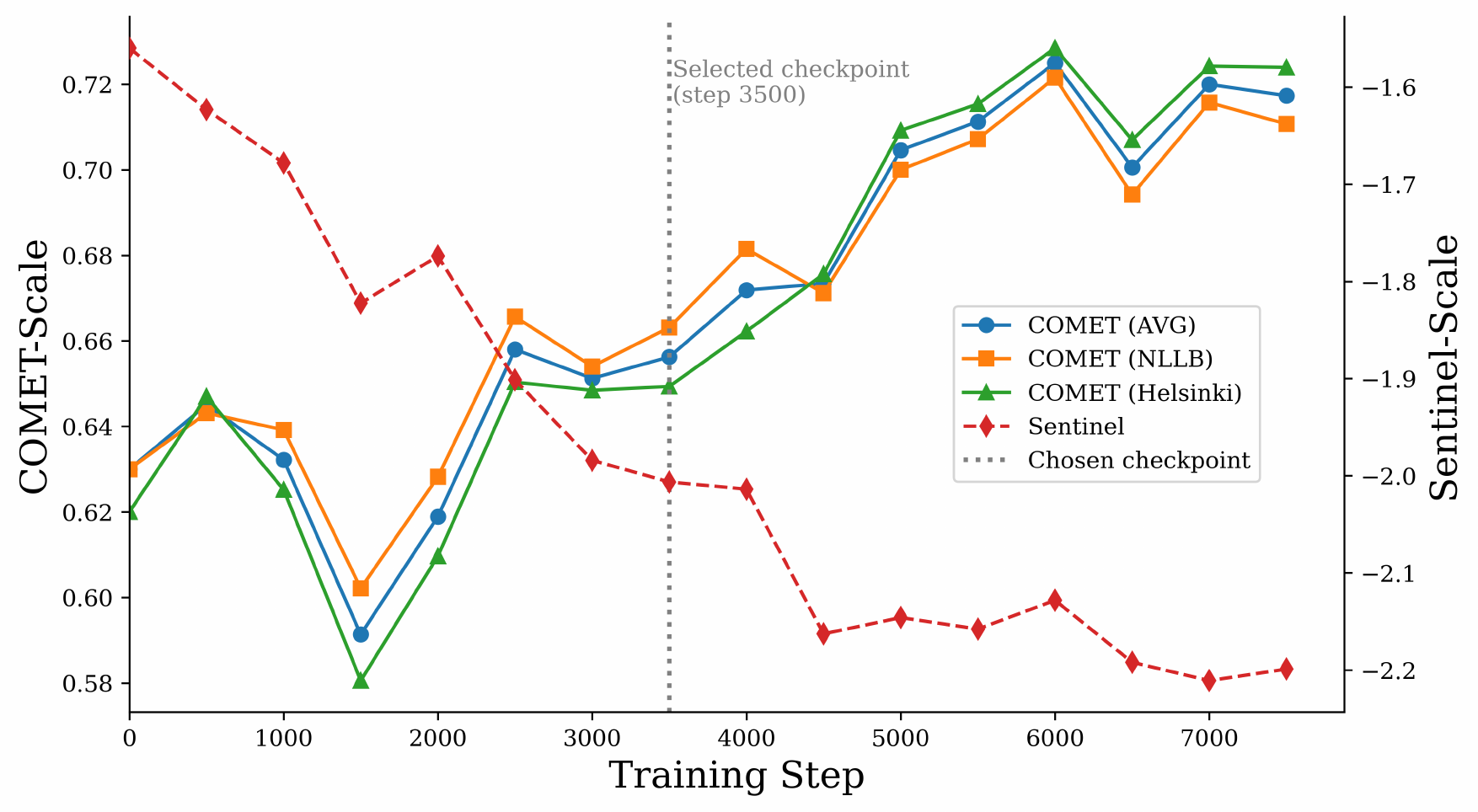}}
\caption{WMT25: Evolution of translation difficulty over training using Sentinel.}
\end{figure}

\section{Sentinel Rewrites}
\label{app:sentinel-rewrites}
This section presents several examples of original instances and their corresponding Sentinel-based rewrites from the selected checkpoint at step 3500. From early training stages onward, the model optimized with Sentinel tends to produce single, very long sentences, in contrast to the COMET-based models, as shown in the examples below.

\begin{examplebox}
\footnotesize
\begin{center}
    \textbf{Original} \\ \textit{Sentinel: -0.76} $|$ \textit{NLLB: 0.41}  $|$ \textit{Helsinki: 0.70}
\end{center}

I am a tweakment holdout. When will wrinkles, bad teeth and big pores be back in fashion?

Totally neutral question, no right or wrong answer: how do you feel about salmon sperm?

\vspace{4pt} \hrule 

\begin{center}
    \textbf{Sentinel} \\ \textit{Sentinel: -1.54} $|$ \textit{NLLB: 0.73}  $|$ \textit{Helsinki: 0.54}\\
\end{center}

I'm a die-hard holdout, while wrinkles, bad teeth, and large pores are due for a retro revival, and I'm utterly indifferent to the sentiment surrounding salmon sperm, how do you feel about it?

\end{examplebox}

\begin{examplebox}
\footnotesize
\begin{center}
    \textbf{Original} \\ \textit{Sentinel: -0.32} $|$ \textit{NLLB: 0.78}  $|$ \textit{Helsinki: 0.72}
\end{center}

I shouldn't be surprised - actual children are buying expensive anti-ageing potions and fretting about their nonexistent crow's feet now. But still, this dramatic normalisation of invasive beauty procedures is unnerving.

\vspace{4pt} \hrule 

\begin{center}
    \textbf{Sentinel} \\ \textit{Sentinel: -0.85} $|$ \textit{NLLB: 0.79}  $|$ \textit{Helsinki: 0.76}\\
\end{center}

I shouldn't be surprised that actual children are indulging in pricey anti-aging elixirs and fretting about their non-existent crow's feet, yet this dramatic normalization of invasive beauty treatments is still unnerving, like a nagging sense that the very notion of youth is being hijacked by a generation that's more concerned with cosmetic camouflage than genuine character.

\end{examplebox}

\begin{examplebox}
\footnotesize
\begin{center}
    \textbf{Original} \\ \textit{Sentinel: 0.30} $|$ \textit{NLLB: 0.84}  $|$ \textit{Helsinki: 0.84}
\end{center}

How the "Manosphere" Became Mainstream Entertainment

\vspace{4pt} \hrule 

\begin{center}
    \textbf{Sentinel} \\ \textit{Sentinel: -0.31} $|$ \textit{NLLB: 0.60}  $|$ \textit{Helsinki: 0.71}\\
\end{center}

The "Manosphere" Mainstreamed Itself into the Fray of Popular Culture.

\end{examplebox}

\begin{examplebox}
\footnotesize
\begin{center}
    \textbf{Original} \\ \textit{Sentinel: -0.62} $|$ \textit{NLLB: 0.58}  $|$ \textit{Helsinki: 0.73}
\end{center}

Bedtime story time!
What’s up with T9 keyboards? Why do 7 and 9 have 4 letters, but others 3. Why those two? Why not assign 1 some letters?
Is 0 always been space?

\vspace{4pt} \hrule 

\begin{center}
    \textbf{Sentinel} \\ \textit{Sentinel: -1.70} $|$ \textit{NLLB: 0.68}  $|$ \textit{Helsinki: 0.22}\\
\end{center}

Bedtime story time, what's the deal with T9 keyboards, where 7 and 9 have four letters, while others have three, and why those two, rather than assigning one or two letters to 1, and has 0 always been relegated to the role of space, a constant in an ever-changing landscape?

\end{examplebox}

\providecommand{\mdl}[1]{{\scriptsize\texttt{#1}}}
\providecommand{\grouprow}[1]{\addlinespace[3pt]\multicolumn{2}{@{}l@{}}{\itshape #1}\\[1pt]}

\section{Models Used}
\label{app:models}

Table~\ref{tab:models} lists models used in this work, together with their role and the checkpoint identifier, to support reproducibility.

\begin{table}[h!]
\centering
\small
\setlength{\tabcolsep}{4pt}
\renewcommand{\arraystretch}{1.15}
\begin{tabular}{@{}>{\raggedright\arraybackslash}p{1.85cm}
                  >{\raggedright\arraybackslash}p{5.0cm}@{}}
\toprule
\textbf{Role} & \textbf{Model / checkpoint} \\
\midrule
\grouprow{Generation and evaluation}
Policy, judge & \mdl{meta-llama/Meta-Llama-3.1-8B-Instruct} \\
\grouprow{Translation difficulty}
QE metric & \mdl{Unbabel/wmt22-cometkiwi-da} \\
MT system 1 & \mdl{facebook/nllb-200-distilled-600M} \\
MT system 2 & \mdl{Helsinki-NLP/opus-mt-en-it} \\
Sentinel & \mdl{sapienzanlp/sentinel-ref-mqm} \\
\grouprow{Constraint rewards}
Embeddings & \mdl{sentence-transformers/all-MiniLM-L6-v2} \\
Acceptability & \mdl{textattack/roberta-base-CoLA} \\
Grammar & \mdl{LanguageTool (en-US)} \\
\grouprow{Text statistics}
Readability & \mdl{spaCy en\_core\_web\_sm} \\
Diversity & \mdl{sacrebleu chrF} \\
\bottomrule
\end{tabular}
\caption{Models used for generation, reward computation, and evaluation. Llama-3.1-8B-Instruct serves
both as the rewriting policy (fine-tuned with LoRA) and, unmodified, as the
LLM judge for naturalness, word rarity, syntax complexity, and topics.
CometKiwi scores the output of both MT systems without a reference, while
Sentinel scores the source alone. The embedding model provides the semantic
similarity reward, computed as cosine similarity over sentence
embeddings.}
\label{tab:models}
\end{table}

\paragraph{Software and versions.}
All experiments use Python 3.10.13, \texttt{torch} 2.10.0 and
\texttt{transformers} 4.57.2. Fine-tuning uses \texttt{trl} 0.23.0
(\texttt{GRPOTrainer}) together with \texttt{unsloth} 2026.3.3 and
\texttt{peft} 0.18.1. Metrics use \texttt{unbabel-comet} 2.2.7,
\texttt{sentence-transformers} 5.2.3, \texttt{language-tool-python} 3.2.2, \texttt{textdescriptives} 2.8.4, \texttt{sacrebleu} 2.6.0 and \texttt{spacy} 3.8.11 (\texttt{en\_core\_web\_sm} 3.8.0). The complete environment is released with the code.

\end{document}